\documentclass{ifacconf}

\usepackage{comment}
\usepackage{tabularx}
\usepackage{booktabs}
\usepackage{caption} 
\usepackage{mathtools}
\usepackage{amsmath}    
\usepackage{amssymb}    
\usepackage{bm}         
\usepackage{url}
\usepackage{graphicx}      
\usepackage{natbib}        
\usepackage{xcolor}
\usepackage{float}      
\usepackage{multirow}
\usepackage{dsfont}
\usepackage{booktabs} 
\usepackage{extdash}
\usepackage[inkscapelatex=false]{svg}
\usepackage{multirow}
\usepackage{array}     
\usepackage{boldline}  

\usepackage{caption}
\begin{document}
\begin{frontmatter}

\title{Enhanced Pruning Strategy for Multi-Component Neural Architectures Using Component-Aware Graph Analysis}

\vspace{-0.2cm}
\thanks[footnoteinfo]{Authors have contributed equally}

\author[1]{Ganesh Sundaram} 
\author[1]{Jonas Ulmen} 
\author[1]{Daniel Görges}
\vspace{-0.2cm}
\address[1]{Department of Electrical and Computer Engineering, \\ RPTU University Kaiserslautern-Landau, Germany \\ 
    (e-mail: \{ganesh.sundaram, jonas.ulmen, daniel.goerges\}@rptu.de).}
\vspace{-0.2cm}

\begin{abstract}

Deep neural networks (DNNs) deliver outstanding performance, but their complexity often prohibits deployment in resource-constrained settings. Comprehensive, structured pruning frameworks based on parameter dependency analysis reduce model size with specific regard to computational performance. When applying them to Multi-Component Neural Architectures (MCNAs), they risk network integrity by removing large parameter groups. We introduce a component-aware pruning strategy, extending dependency graphs to isolate individual components and inter-component flows. This creates smaller, targeted pruning groups that conserve functional integrity. Demonstrated effectively on a control task, our approach achieves greater sparsity and reduced performance degradation, opening a path for efficiently optimizing complex, multi-component DNNs.

\end{abstract}


\end{frontmatter}

\section{Introduction}

While deep learning's tremendous advances yield increasingly complex and high-performing neural networks, especially for visual data, a critical conflict emerges. The very attributes driving this success—large model sizes, demanding computations, and intensive matrix operations—clash directly with the limitations of consumer electronics and edge devices such as microcontrollers, hindering practical deployment where resources are scarce.

Given these challenges, reducing model complexity while preserving performance is essential. Model compression techniques such as \emph{Pruning}, \emph{Quantization}, and \emph{Knowledge distillation} have become essential. Pruning is undoubtedly effective in practice, and there exist sound theoretical hypotheses on how a pruned network might yield similar performance compared to its full-sized counterpart \citep{frankleLotteryTicketHypothesis2019}.

In architectures where different network modules serve different purposes, e.g., encoders, prediction models, and control policies, as in TD-MPC \citep{hansen2022temporaldifferencelearningmodel}  or JEPA \citep{assran2023selfsupervisedlearningimagesjointembedding, ulmen2025learning},  dependency graphs, which are commonly used for pruning, cannot capture the intricate data flows between the components. Unlike monolithic networks, such Multi-Component Neural Architectures (MCNA) demand a more granular dependency model to represent the intra- and inter-component interactions accurately. Additionally, although all components of MCNAs have to be trained, not all components are used in downstream tasks, which should be accounted for when prioritizing which component to prune.

Two common pruning libraries are \textsf{prune} from \texttt{PyTorch}  and \textsf{Torch-Pruning} \citep{fangDepGraphAnyStructural2023f}. The former masks weights without reducing the parameter count, limiting its effectiveness in shrinking model size. While \textsf{Torch-Pruning} uses a dependency graph to determine pruning groups, it struggles with MCNAs by creating large groups that span multiple components, leading to significant performance degradation. These limitations are discussed further in detail in later sections.

In this paper, we propose an extended dependency graph approach for structured pruning that explicitly isolates individual components and captures inter-component data flows in MCNAs. By generating smaller, targeted pruning groups, our method preserves the overall network integrity while substantially reducing complexity. We validate our approach on a control task using a trained TD-MPC model, demonstrating increased sparsity with less performance degradation.

\section{Related Work}

Simple magnitude-based pruning, which removes weights with the smallest absolute values, was one of the earliest techniques. Despite its simplicity, this approach has shown surprising effectiveness, often matching or outperforming more complex methods \citep{guptaComplexityRequiredNeural2022}. %
%
However, the field has expanded to include structured pruning. Structured pruning removes entire sets of parameters, enabling direct computational and memory savings. This approach has become particularly important for deploying models on edge devices \citep{heStructuredPruningDeep2023}. Structurally Prune Anything (SPA) has been developed to prune any architecture from any framework and at any stage of training \citep{wangStructurallyPruneAnything2024}. %

Recent advancements in pruning offer more sophisticated parameter evaluation through Hessian-based and class-aware methods \citep{chongResourceEfficientNeural2023, jiangClassAwarePruningEfficient2023}. To counteract potential accuracy degradation from pruning, techniques like knowledge distillation are employed to boost performance \citep{qianBoostingPrunedNetworks2024}. Concurrently, sparse optimization techniques integrate regularization during pretraining and pruning, achieving competitive results without needing subsequent fine-tuning \citep{shiSparseOptimizationGuided2024b}. Theoretical progress is also evident, with operator-theoretic perspectives unifying magnitude and gradient-based pruning \citep{redmanOperatorTheoreticView2022}. Pruning is further integrated with gradient information \citep{yangDecayPruningMethod2024} and quantization, enabling automatic trade-offs for hardware efficiency \citep{wangDifferentiableJointPruning2020}. Finally, reflecting the growing emphasis on sustainable AI \citep{chengSurveyDeepNeural2024b}, energy-constrained optimization methods explicitly target energy reduction by formulating pruning under energy budget constraints \citep{liuEnergyConstrainedOptimizationBasedStructured2022}. 
These methods - and to our knowledge, no existing method - try to formalize pruning concerning the functional and structural interdependence of different network components.

Structured pruning's hardware compatibility facilitates real-time inference, aided by compiler optimizations and fine-grained methods \citep{gongAutomaticMappingBestSuited2021}. Frameworks like HALP globally optimize accuracy under latency constraints \citep{shenStructuralPruningLatencySaliency2022}. Demonstrated speedups include 2x for YOLOv7 (FPGA) \citep{pavlitskaIterativeFilterPruning2024}, 19.1x for LSTM (Jetson) \citep{lindmarIntrinsicSparseLSTM2022}, and 2x for GNNs (PruneGNN on A100) \citep{gurevinPruneGNNAlgorithmArchitecturePruning2024b}. It enhances compute-in-memory (11.1x compression) \citep{mengExploringComputeinMemoryArchitecture2022b} and systolic arrays (SCRA, 4.79x speedup) \citep{zhangSCRASystolicFriendlyDNN2023b}. Hardware-architecture co-design approaches like CSP further improve efficiency by preserving data reuse and utilizing predictable sparsity \citep{hansonCascadingStructuredPruning2022}.

One of the most significant advancements in structured pruning is the development of generalizable frameworks that can be applied across various neural network architectures. Traditional pruning methods often rely on manually designed grouping schemes, which are non-generalizable to new architectures. DepGraph is a general and automatic structured pruning framework that overcomes the limitations of traditional methods. Although using simple norm-based criteria, DepGraph's versatility and ability to ensure consistent pruning lead to strong performance across various domains and architectures, confirmed through extensive evaluations \citep{fangDepGraphAnyStructural2023f}. Since DepGraph is the basis of this work, we will further analyze and critique the framework in the remaining sections.

\section{Model Compression and Its Challenges}

\subsection{Model Compression for Neural Network Controllers: Challenges Beyond Standard MCNAs}

Not all demands on specific neural network components are created equal, leading to vastly different compression challenges. For models predominantly utilized in perception or language tasks, compression primarily involves optimizing the trade-off between model size reduction and the preservation of predictive accuracy. Conversely, learned control systems, particularly those deployed within physical processes such as robotics, impose substantially stricter constraints. In this context, compression must not only address efficiency but critically ensure adherence to strict control-theoretic requirements—including stability, safety, and robustness—which are imperative for maintaining the technical integrity and economic viability of the application.


In the latter case, even minimal component-agnostic pruning can disrupt system behavior so that the imposed requirements are no longer met. A fundamental relationship governing component interaction is their functional sequence within the network's execution flow, where each step relies on the sufficient performance of its predecessor.
For example, in the aforementioned architectures, TD-MPC and JEPA, a slight alteration of the encoder's parameters can severely deform the learned manifold and adversely affect the succeeding world model by providing an unfamiliar latent representation. This interdependency underscores that downstream module fidelity critically depends on the output quality of upstream components. Thus, effective compression must extend beyond mere structural simplification.

\subsection{Limitation of Existing Dependency Modeling}

\emph{DepGraph}, introduced in \citep{fangDepGraphAnyStructural2023f}, captures inter-layer dependencies in structured pruning by representing each layer as a node and drawing edges when pruning one layer forces adjustments in another, such as when removing a convolution channel necessitates pruning its associated batch normalization or subsequent convolution layers. The method distinguishes between \emph{intra-layer} and \emph{inter-layer} dependencies, effectively working for monolithic networks across CNNs, transformers, RNNs, and GNNs.

Unfortunately, \emph{DepGraph} treats the entire MCNA as a single chain of layers, merging connected nodes into coarse groups. This leads to two main issues: first, neglecting module boundaries results in overly large dependency groups, where even minimal pruning can severely impact performance. Second, when such groups span multiple components, they fail to account for the varying sensitivity of individual components, potentially compromising downstream performance.

\subsection{Further Challenges in MCNA}
\label{subsection:otherMCNAComp}

In complex MCNAs, module interactions extend far beyond simple feedforward links, often incorporating feedback loops, recurrent connections, and dynamic gating. Standard \emph{DepGraph}, which relies on static, direct dependencies, oversimplifies these relationships. Feedback loops, for example, can either obscure key dependencies or cause multiple layers to merge into oversized groups. Recurrent connections—such as planning modules that repeatedly invoke a dynamics model—might not be properly captured, misaligning the temporal dependencies critical for proper function. Likewise, gating and conditional execution introduce dynamic connectivity that \emph{DepGraph} fails to distinguish. These limitations lead \emph{DepGraph} to treat all components equally, neglecting their unique functionalities, sensitivities, and interdependencies, and ultimately compromising pruning effectiveness. 

Although dependency graph-based pruning is a valuable foundation, the modular but interlinked nature of MCNAs requires a more nuanced strategy. To address this limitation, we propose an extension to \emph{DepGraph} that explicitly models component boundaries and inter-component flows, capturing sensitive dependencies to ensure that pruning preserves both structural and functional integrity.


\section{Constructing Dependency Graphs for MCNAs}

Using the component-aware grouping method, the network architecture is partitioned into \(N\) components. For any two layers \(i\) and \(j\) (which may belong to any of these \(N\) components), we define the overall dependency \(D(f_i^-, f_j^+)\) by combining both intra-component and inter-component dependencies.

\subsection{Extending the Dependency Criterion}

For clarity, we introduce the notation used in this section: 
\begin{itemize}
    \item \(f_k^-, f_k^+\): input and output tensors of layer \(k\)
    \item \(C_k\): component to which layer \(k\) belongs, with \(C_k \in \{C_1, C_2, \dots, C_N\}\)
    \item \(C_k^-, C_k^+\): designated input/output tensors or interfaces of component \(C_k\)
    \item \(\leftrightarrow\): denotes a verified connection (data flow)
    \item \(\operatorname{sch}(f)\): specifies the pruning scheme applied to the feature tensor \(f\)
    \item \(\mathds{1}[\cdot]\): indicator function (1 if true, 0 otherwise)
    \item Logical operators \(\land\) and \(\lor\) denote AND and OR, respectively
\end{itemize}

\subsection{Intra-Component Dependency (\(D_{I}\))}

This term applies when both layers belong to the same component (i.e., \(C_i = C_j\)). It captures two aspects: \emph{inter-layer} data flow
\begin{equation}
    d_1(i, j) = \mathds{1}[f_i^- \leftrightarrow f_j^+]
\end{equation}
(which is true if the output of layer \(j\) feeds the input of layer \(i\)), and \emph{intra-layer} dependency
\begin{equation}
    d_2(i, j) = \mathds{1}[i = j \land \operatorname{sch}(f_i^-) = \operatorname{sch}(f_j^+)]
\end{equation}
(enforcing that a layer’s input and output adhere to a consistent pruning scheme).

Thus, the intra-component dependency is summarized by
\begin{equation}
    D_{I}(i, j) = \mathds{1}[C_i = C_j] \land \Bigl( d_1(i, j) \lor d_2(i, j) \Bigr).
    \label{eq:dep_intra_short_no_brace}
\end{equation}

\textit{Explanation}: \(D_I(i, j)\) is true if layers \(i\) and \(j\) are in the same component (\(\mathds{1}[C_i = C_j]\)) and either the output of layer \(j\) directly feeds the input of layer \(i\) (\(\mathds{1}[f_i^- \leftrightarrow f_j^+]\)) or they represent the same layer (\(i=j\)) and share a coupled pruning scheme (\(\operatorname{sch}(f_i^-) = \operatorname{sch}(f_j^+)\)).

\subsection{Inter-Component Interface Dependency (\(D_{X}\))}

This term addresses dependencies between layers residing in different components (i.e., \(C_i \neq C_j\)). It is activated when layer \(j\) produces the designated output interface of component \(C_i\) and layer \(i\) consumes the input interface of component \(C_j\). There exists a verified connection between the interfaces (\(C_i^+ \leftrightarrow C_j^-\)).

Formally, we write:
\begin{align}
    D_{X}(i, j) = \mathds{1}[C_i \neq C_j] & \land 
 d_1(i, j)
    \label{eq:dep_inter_short_no_brace}
\end{align}

\textit{Explanation}: \(D_X(i, j)\) is true if layers \(i\) and \(j\) are in different components (\(\mathds{1}[C_i \neq C_j]\)) and layer \(j\) generates the output interface of \(C_i\), while layer \(i\) consumes the input interface of \(C_j\), with a verified data flow existing from \(C_i\)'s output to \(C_j\)'s input.

\subsection{Overall Dependency (\(D\))} 

The complete dependency function is defined as
\begin{align}
    D(f_i^-, f_j^+) = D_{I}(i, j) \lor D_{X}(i, j)
    \label{eq:dep_final_assembled_short_revised}
\end{align}
and is evaluated for every pair of layers \(i\) and \(j\) in the network. This formulation separates the logic for handling dependencies within components (\(D_I\)) from that at the interfaces between components (\(D_X\)) and spans all \(N\) components. In doing so, it ensures that both intra- and inter-component dependencies are properly captured for structured pruning, even in architectures where components are loosely coupled but structurally interdependent.

\begin{figure*}[ht]
    \centering
    \includegraphics[width=\textwidth]{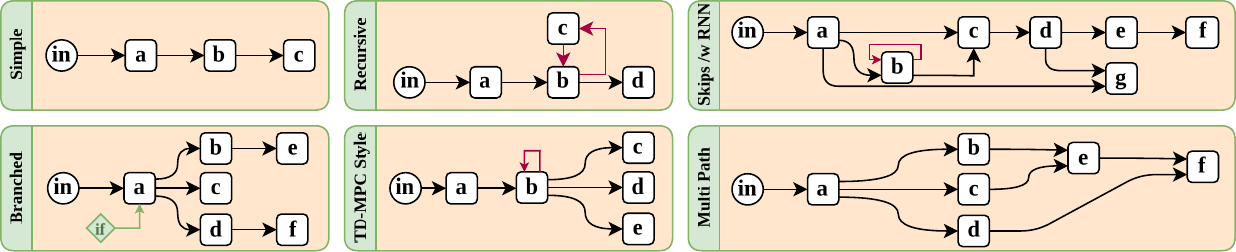} 
    \caption{Schematics of various MCNA architectures of increasing complexity—from simple to branched and recursive forms, selected to evaluate the performance of the component identification and inter-dependency detection module in our component-aware approach. Recurrent layers are represented by arrows colored in red, while conditional variables are indicated by arrows colored in green. Detailed information on each component is provided in Table \ref{tab:metricsnew}.}

    \label{fig:MCNAmodels}
\end{figure*}

\subsection{Identifying Components and Their Dependencies}
\label{subsec:intercompdep}

We extract discrete components from an MCNA by leveraging the hierarchical structure of \texttt{PyTorch's} \texttt{nn.Module}. Initially, \texttt{model.named\_children} is used to identify the high-level modules that constitute the network. For finer granularity, \texttt{model.\allowbreak named\_modules} is applied recursively to capture all architectural elements, thereby defining our component set \( C = \{C_1, C_2, \ldots, C_n\} \). To uncover inter-component dependencies, we first construct a candidate dependency graph assuming every possible connection, yielding $n^{2}$ directed edges (including self-loops). We then refine this graph by eliminating the relations that are not utilized in the actual forward pass through the whole network. 

Dynamic tracing is performed using \texttt{PyTorch} \textsf{forward} and \textsf{backward hooks}. The \textsf{forward hooks} record the runtime inputs and outputs along their dimensions of individual components, while \textsf{backward hooks} capture gradient flows to make sure that the structural integrity is not lost and further training is still possible. The forward and backward flows are the basis for adjustments to the dependency edges.

The dependency graph obtained can be further verified through certain techniques. \emph{Shape matching} ensures that the output dimension from a component must match the corresponding receiver’s input dimension, and if multiple components contribute to an input, their combined output dimensions should equal the total input dimension of the receiving component. These rule out any dimensional discrepancies that suggest missing or erroneous dependencies. \emph{Dynamic tensor tracing} via the \texttt{PyTorch} \textsf{autograd} graph records the lineage of intermediate tensors during each forward pass. In cases of non-standard data manipulations (e.g., reshaping, slicing, aggregating) or when components share buffers or weights, \emph{semantic heuristics} instantiate plausible connections again based on the network's intended functionality.

\subsection{Pruning Groups and Importance Calculation}
\label{subsec:pruninggroups}

After identifying individual components and their interdependencies, we construct a dependency graph for each component and extract corresponding pruning groups. In addition, output-input dependencies that span components are isolated as separate groups. For cyclic dependencies arising from feedback loops or recurrent connections, we build specialized subgraphs that accurately capture the functional succession and conditional relationships within each component.

Isolating pruning groups on a per-component basis confines the scope of pruning decisions, preventing an infeasible change of the dimensions between connected components. Additionally, we enable the expert user to exclude certain components from (further) pruning, e.g., to prevent further removal of parameters in an upstream component, so that performance degradation does not cascade to downstream components. For each group, we compute an importance metric, utilizing $\ell_1$ or $\ell_2$ norms of the weights to determine the order of pruning. 

\section{Experiments and Discussion}

\begin{table*}[!t]
\caption{The table provides an overview of the evaluation results for the models depicted in Figure \ref{fig:MCNAmodels}. It lists the layer input-output dimensions for each component group in the format \(\langle\mathbf{component\ name}: (depth,\; input\ dim \to output\ dim)\rangle\). Furthermore, it displays both the total count of groups and the average group size (operations per group) as determined by the vanilla dependency graph method versus our component-aware approach, in addition to the inter-component groups found using the latter.}

\label{tab:metricsnew}
\centering
\normalsize
\setlength{\tabcolsep}{1.8pt}             
\renewcommand{\arraystretch}{0.85}       
\begin{tabular}{>{\raggedright\arraybackslash}p{1.5cm} c >{\raggedright\arraybackslash}p{6cm} c c c c c}
\toprule
Model & \# Comp. & \multicolumn{1}{c}{Input-Output Layers Dim.} & \multicolumn{2}{c}{Total Groups} & \multicolumn{2}{c}{Avg. Group Size} & Cross-Comp. \\
                                                                    \cmidrule(lr){4-5}\cmidrule(lr){6-7}
      &          &                                              & Vanilla & Comp-Aware & Vanilla & Comp-Aware & Groups \\
\midrule

Simple 
& 3 
& \begin{tabular}{@{}p{2.9cm} p{2.9cm}@{}}
$\langle\mathbf{a}: (1,\;128\to20)\rangle$ & $\langle\mathbf{b}: (1,\;20\to15)\rangle$ \\[1mm]
$\langle\mathbf{c}: (1,\;15\to1)\rangle$ & 
\end{tabular}
& 6  & 8  & 4.83 & 1.63 & 2 \\
\midrule

Branched 
& 6 
& \begin{tabular}{@{}p{2.9cm} p{2.9cm}@{}}
$\langle\mathbf{a}: (2,\;784\to64)\rangle$ & $\langle\mathbf{b}: (2,\;64\to48)\rangle$ \\[1mm]
$\langle\mathbf{c}: (2,\;64\to10)\rangle$ & $\langle\mathbf{d}: (2,\;64\to96)\rangle$ \\[1mm]
$\langle\mathbf{e}: (1,\;48\to392)\rangle$ & $\langle\mathbf{f}: (1,\;96\to784)\rangle$
\end{tabular}
& 20 & 33 & 4.10 & 1.73 & 3 \\
\midrule

Multi-Path 
& 7 
& \begin{tabular}{@{}p{2.9cm} p{2.9cm}@{}}
$\langle\mathbf{a}: (4,\;784\to64)\rangle$ & $\langle\mathbf{b}: (2,\;64\to32)\rangle$ \\[1mm]
$\langle\mathbf{c}: (3,\;64\to32)\rangle$ & $\langle\mathbf{d}: (1,\;64\to32)\rangle$ \\[1mm]
$\langle\mathbf{e}: (3,\;64\to32)\rangle$ & $\langle\mathbf{f}: (2,\;64\to64)\rangle$ \\[1mm]
$\langle\mathbf{g}: (3,\;64\to5)\rangle$ & 
\end{tabular}
& 32 & 46 & 4.25 & 1.87 & 6 \\
\midrule

Recursive 
& 4 
& \begin{tabular}{@{}p{2.9cm} p{2.9cm}@{}}
$\langle\mathbf{a}: (4,\;64\to5)\rangle$ & $\langle\mathbf{b}: (3,\;10\to64)\rangle$ \\[1mm]
$\langle\mathbf{c}: (1,\;64\to64)\rangle$ & $\langle\mathbf{d}: (2,\;64\to64)\rangle$
\end{tabular}
& 10 & 24 & 4.47 & 2.00 & 3 \\
\midrule

TDMPC-Style 
& 4 
& \begin{tabular}{@{}p{2.9cm} p{3.1cm}@{}}
$\langle\mathbf{a}: (4,\;784\to32)\rangle$ & $\langle\mathbf{b}: (3,\;32\to4)\rangle$ \\[1mm]
$\langle\mathbf{c}: (5,\;36\to32)\rangle$ & $\langle\mathbf{d}: (4(2),\;36\to1)\rangle$
\end{tabular}
& 28 & 39 & 4.57 & 2.26 & 2 \\
\midrule

Complex CNN 
& 7 
& \begin{tabular}{@{}p{2.9cm} p{2.9cm}@{}}
$\langle\mathbf{a}: (1,\;72\to64)\rangle$ & $\langle\mathbf{b}: (1,\;64\to64)\rangle$ \\[1mm]
$\langle\mathbf{c}: (7,\;784\to64)\rangle$ & $\langle\mathbf{d}: (2,\;64\to64)\rangle$ \\[1mm]
$\langle\mathbf{e}: (1,\;64\to64)\rangle$ & $\langle\mathbf{f}: (4,\;64\to2)\rangle$ \\[1mm]
$\langle\mathbf{g}: (2,\;64\to1)\rangle$ &
\end{tabular}
& 36 & 52 & 5.80 & 2.04 & 7 \\
\bottomrule
\bottomrule
\end{tabular}
\end{table*}

Our extended dependency graph for MCNAs was designed to form smaller, more suitable groups to minimize the adverse effects of pruning. This was achieved by modeling the per-component dependency and combining them into a dependency graph for the MCNA. In our first experiment, we created various MCNA architectures of increasing complexity, from simple to branched and recursive forms as shown in Figure~\ref{fig:MCNAmodels}, and evaluated our method's component identification. As detailed in Table~\ref{tab:metricsnew}, our approach successfully identified the number of components as well as the input-output layer dimensions for each component group. For example, the \textit{Simple} model was decomposed into three components with clearly defined dimensions, while more complex architectures, such as the \textit{Branched} and \textit{Multi-Path} models, were appropriately segmented, reflecting their inherent structural complexity.

In the second experiment, we investigated our grouping approach by comparing the vanilla dependency graph method with our component awareness. The vanilla method produced fewer total groups with larger average group sizes, indicating coarser groupings. In contrast, our method generated a higher number of groups with smaller sizes; for instance, in the \textit{Branched} model, the vanilla method resulted in 20 groups with an average group size of 4.10 operations, whereas our component-aware method produced 33 groups with an average size of 1.73 operations. This demonstrates that our approach yields finer, more precise partitions of the architecture.

\begin{figure}[htbp] 
  \centering
    \centering
    \includegraphics[width=1\linewidth]{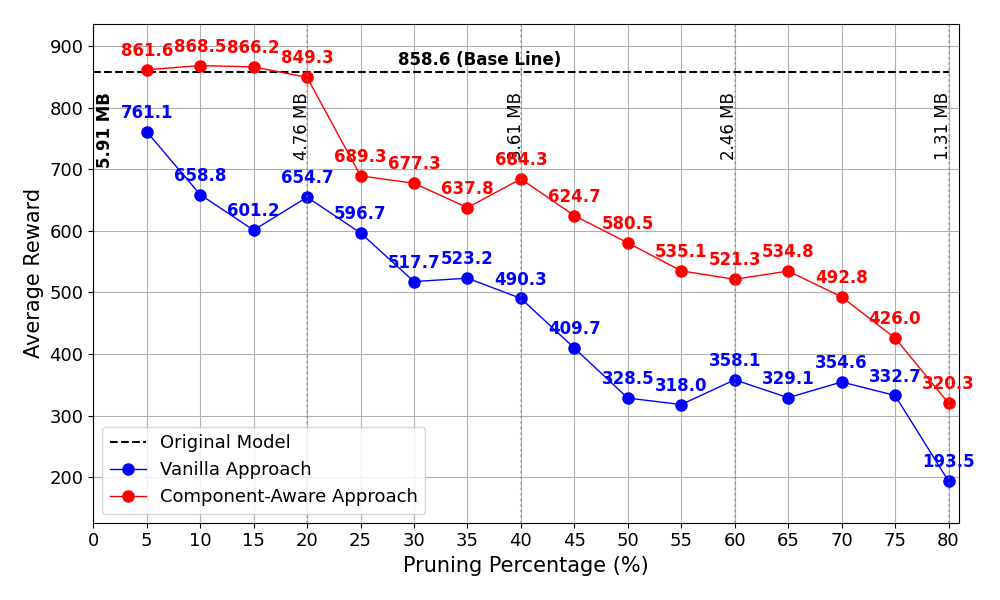}
    \label{fig:TD-MPCpruningresult_left}
    \caption{Average reward for a TD-MPC model under varying pruning percentages, comparing the vanilla pruning approach (blue) and the component-aware approach (red). The unpruned baseline reward is indicated in black. The figure also shows the approximate total model size (in MB) at each pruning level, providing a reference for how pruning reduces the overall network footprint.}

  \label{fig:TD-MPCpruningresult}
\end{figure}

To further assess whether creating finer groups translates into tangible benefits in both performance preservation and model size reduction, we conducted a final experiment using a TD-MPC model on the cartpole-swingup Gym environment with image inputs \citep{brockman2016openaigym}. An over-parameterized TD-MPC model (with each component having a width of 512 and a depth of three layers) was first trained until the average reward stabilized around 850. To evaluate its effectiveness, we compared our component-aware method against the baseline (vanilla dependency graph) approach across various sparsity levels, from 5\% to 80\%. The results, presented in Figure~\ref{fig:TD-MPCpruningresult}, reveal a sheer difference in performance degradation. The baseline method suffered an immediate performance drop, culminating in a catastrophic 50\% reward loss at just 40\% sparsity. In contrast, our component-aware approach exhibited a much more graceful and gradual decline, demonstrating its superior robustness.

Moreover, by revealing the constituent components within each pruning group, our method offers the flexibility to protect sensitive components from being overly pruned. This was demonstrated by assigning higher importance weighting to groups associated with the encoder in the TD-MPC model, thereby reducing the likelihood of excessive pruning in these critical regions. Overall, these results substantiate that our component-aware approach not only provides a more granular grouping of the MCNA architectures but also preserves performance more effectively, even during aggressive pruning. This represents a significant advancement over existing methods, enabling model size reduction without a substantial loss in performance.

\section{Conclusion}

In this work, we demonstrated that switching to a component-aware dependency graph produces smaller, targeted, structured pruning groups while conserving functional integrity. Our method, which identifies network components and accurately models both intra- and inter-dependencies, leads to reduced average group size and thereby mitigates the overall pruning impact. Because we ran all inference on a standard PC in \texttt{PyTorch}’s eager mode, we observed only modest speed‑ups as pruning increased; in future work, we will explore inference‑time gains using embedded hardware and optimized inference engines.

Our ongoing research is focused on expanding the importance criteria of these groups by incorporating sensitivity analyses and theoretical constraints, including stability guarantees. Future work will explore post-pruning fine-tuning to enhance performance further, as well as the integration of component-aware pruning with other model reduction techniques like \emph{quantization}. To conclusively demonstrate the method's impact and scalability, we will then broaden our trials to more complex architectures, such as Large Language Models (LLMs), and a diverse range of applications.

\bibliography{ifacconf}             

\end{document}